\documentclass[11pt]{article}

\usepackage[letterpaper,margin=1in]{geometry}
\usepackage{iftex}
\usepackage{graphicx}
\usepackage{booktabs}
\usepackage{amsmath,amssymb}
\usepackage{cite}
\usepackage{authblk}
\usepackage{tabularx,array}
\usepackage[hidelinks]{hyperref}
\usepackage{microtype}

\AtBeginDocument{%
  }

\ifPDFTeX
  \PackageError{manuscript}{Khmer script requires XeLaTeX or LuaLaTeX}{Compile this manuscript with XeLaTeX (recommended) or LuaLaTeX.}
\else
  \usepackage{fontspec}
  \newfontfamily\khmerfont[Script=Khmer,Scale=MatchLowercase,Path=./]{NotoSansKhmer.ttf}
\fi

\providecommand{\Description}[1]{}
\newcolumntype{Y}{>{\raggedright\arraybackslash}X}

\begin{document}
\title{Evaluation of Chunking Strategies for Effective Text Embedding in Low-Resource Language on Agricultural Documents}

\author[1]{Sovandara Chhoun}
\author[2]{Pichdara Po}
\author[2]{Sereiwathna Ros}
\author[3]{Wan-Sup Cho}
\author[3]{Saksonita Khoeurn}

\affil[1]{Department of Big Data, Chungbuk National University, Cheongju-si, South Korea\\
}
\affil[2]{Department of Computer Science, Chungbuk National University, Cheongju-si, South Korea\\
}
\affil[3]{BigDataLabs Co., Ltd. Department of Management Information Systems, Chungbuk National University, South Korea\\
}
\date{}
\maketitle
\vspace{-3.0em}
\begin{abstract}
In this study, we compare the performance of four text chunking approaches — Recursive, Khmer-Aware, Sentence-Based, and LLM-Based — within a Retrieval-Augmented Generation (RAG) framework applied to Khmer agricultural documents. The document chunks are encoded using the BGE-M3 multilingual embedding model and retrieved using the FAISS library. Performance is evaluated using four metrics: Average Retrieval Score (L2 distance), Answer Relevance, Khmer Coverage, and Khmer Intersection over Union, all measured against ground-truth question–answer pairs. For evaluation, we perform 5-fold cross-validation over 18 question–answer pairs. We observe the best performance for the character-based Recursive chunking method with a chunk size of 300 characters, achieving the lowest L2 distance (0.4295 ± 0.0461), highest Answer Relevance (0.8663 ± 0.0199), and highest Khmer IoU (0.6441 ± 0.0347). A paired t-test shows a statistically significant improvement over the Sentence-Based chunking method in L2 distance (p = 0.0121). These results highlight the importance of segmentation granularity and structural preservation for optimizing dense retrieval in morphologically complex, low-resource languages such as Khmer.
\end{abstract}

\section{Introduction}
Recent breakthroughs in embedding-based natural language processing have revolutionized the approach to handling textual data, including indexing, retrieval, and semantic analysis within modern intelligent systems \cite{lewis2020rag}. The dense vector representations learned via deep neural embedding architectures facilitate semantic similarity search, document retrieval, and question-answering beyond simple word-level matching \cite{chen2024bge}. The standard pipeline involves document preprocessing, chunking, embedding generation, and vector-based indexing using approximate nearest neighbor search libraries such as FAISS \cite{johnson2017faiss}. Although significant research has improved embedding architectures and retrieval paradigms, there has been relatively little work on improving the document chunking step, which is essential for dividing long documents into semantically coherent chunks before embedding \cite{qu2024semanticchunking}.

Chunking is arguably the most significant step in document preparation since it directly influences the granularity and semantic content of embedded document representations \cite{reimers2019sentencebert, gunther2025latechunking}. Poor document segmentation can result in embeddings being too generic, combining unrelated concepts, or too fine-grained, disrupting semantic relationships between concepts and procedures. Hence, document chunking is an understudied yet vital component in embedding-based document retrieval and semantic search systems \cite{lewis2020rag, guu2020realm}. Despite its importance, there has been little work on evaluating document chunking strategies, especially in languages with limited resources \cite{chea2015khmer}.

These issues are particularly exacerbated when dealing with low-resource languages, where linguistic properties make preprocessing and modeling more difficult \cite{chea2015khmer, buoy2020khmer}. Khmer is an interesting case study as it is an official language of Cambodia with no word boundaries, complex orthography, and limited downstream natural language processing (NLP) resources \cite{sry2022khmerreview}. Traditional chunking methods, such as fixed window-based or sentence-based chunking, have been developed with reference to high-resource languages with well-established punctuation and grammatical rules \cite{koshorek2018textsegmentation, alemi2015textsegmentation}. Applying such methods to Khmer may lead to a loss of semantic continuity or even damage procedural content. This is particularly critical when dealing with agricultural documents that often require multi-step procedural content, specialized vocabulary, and hierarchical explanatory content \cite{chea2015khmer, buoy2020khmer}. Maintaining semantic continuity is critical to ensure accurate retrieval and question answering \cite{lewis2020rag, guu2020realm}.

Inspired by this problem, this study aims to investigate how chunking strategy selection affects embedding-based retrieval performance on Khmer agricultural documents. More precisely, this study aims to answer the following research questions:

\begin{itemize}
  \item \textbf{1)} How do different chunking strategies affect retrieval accuracy in Khmer agricultural documents?
  \item \textbf{2)} What chunk size optimizes semantic coherence and retrieval precision within a recursive segmentation framework?
  \item \textbf{3)} What are the computational trade-offs between rule-based and LLM-based chunking approaches?
\end{itemize}

To conduct this investigation, we carry out an extensive evaluation of four different chunking methods: recursive chunking, Khmer-aware chunking, sentence-based chunking, and large language model chunking. The methods we employ are based on various segmentation philosophies, which can be broadly categorized as rule-based segmentation and semantic segmentation \cite{alemi2015textsegmentation}.

For embedding and retrieval, we utilize the multilingual model BGE-M3 \cite{chen2024bge}, whose vector representations are indexed using FAISS \cite{johnson2017faiss}. The evaluation metrics are based on multiple criteria, including Average Retrieval Score (L2 Distance), Answer Relevance (Cosine Similarity), Khmer Coverage, and Khmer Intersection-over-Union (Khmer IoU) \cite{sry2022khmerreview}. We adopt a 5-fold cross-validation approach to ensure statistical robustness in our evaluation.

\section{Related Works}
\subsection{Text Chunking in Embedding-Based Retrieval}

Text chunking is a key component in embedding-based information retrieval systems, as it determines the semantic granularity of the content being retrieved. In text-based information retrieval systems that incorporate a retrieval-augmented generation (RAG) mechanism, text segmentation is a key component that influences the accuracy of document embeddings and, in turn, information retrieval accuracy \cite{lewis2020rag, guu2020realm}. Text chunking is a technique that splits long documents into smaller semantic units that can easily be embedded and processed in large language models \cite{merola2025reconstructing}. As a result, the text segmentation technique significantly determines the accuracy of document embeddings in representing contextual information in the source text.

Previous techniques in text-based information retrieval systems relied on fixed-size text chunking, in which a text document is segmented into smaller chunks of equal character or token size, usually with overlapping segments to preserve contextual information. Although this technique is efficient in text processing and information retrieval, it is prone to information fragmentation, in which semantically significant information is spread across multiple chunks or irrelevant information is embedded in a chunk \cite{reimers2019sentencebert}. Large chunk sizes may result in a lack of semantic specificity, while small chunk sizes may fragment contextual information in a text document. Recent studies show that recursive text segmentation techniques can significantly improve information retrieval accuracy compared to fixed-size text segmentation techniques \cite{qu2024semanticchunking, gunther2025latechunking}. However, there is a lack of systematic evaluations of various text segmentation techniques \cite{koshorek2018textsegmentation, alemi2015textsegmentation}.
\subsection{Sentence-Based and Language-Aware Chunking}

Sentence-based chunking methods are designed to maintain semantic integrity by dividing the text according to sentence boundaries. More advanced semantic chunking methods involve computing embeddings for adjacent groups of sentences and identifying topic shifts based on semantic distance thresholds \cite{qu2024semanticchunking}. This method is likely to yield optimal results for resource-rich languages for which sentence tokenizers are robust and punctuation is standardized.

Sentence-based chunking methods, however, are likely to face challenges in low-resource languages, where punctuation may be irregular and robust NLP tools may not be available. Language-aware preprocessing methods have been developed to overcome these challenges. For example, in the case of the Vietnamese language, NLP methods such as VnCoreNLP use conditional random fields (CRF) and neural sequence models to improve phrase segmentation \cite{vu2018vncorenlp}. Similar methods have been developed for the Khmer language, including CRF-based word segmentation \cite{chea2015khmer}, segmentation methods based on BiLSTM networks \cite{buoy2020khmer}, and surveys of word segmentation methods for the Khmer language \cite{sry2022khmerreview}. Despite these studies highlighting the importance of language-aware segmentation, their effect on the performance of embedding-based retrieval systems has not been extensively explored \cite{seahelm2025, bhasa2023}.

\subsection{LLM-Based Document Segmentation}

Recently, large language models (LLMs) have been utilized for document segmentation using prompt-based methods. Contextual retrieval-based methods use the entire document to prompt an LLM to generate enriched descriptions for document segments before embedding, as shown in previous works \cite{Strauss_2025}. On the other hand, LLM-based semantic chunking methods prompt an LLM to split documents based on semantic boundaries inferred by the model.

While these methods provide flexibility for different languages and domains, they come at the cost of increased computational latency. Contextual retrieval-based methods, such as the combination of dense embeddings and sparse retrieval, have been shown to yield significant reductions in retrieval errors \cite{Strauss_2025}. However, the assessment of LLM-based document chunking strategies has been inadequate for low-resource languages such as Khmer \cite{seahelm2025, bhasa2023}.

\subsection{Embedding Models for Multilingual Retrieval}

Dense models of embeddings allow text segments to be mapped to a continuous vector space, facilitating semantic similarity measurement using cosine similarity or L2 distance. Multilingual embeddings are an extension of this concept that aims to perform this operation even in low-resource languages. Embeddings such as mBERT, LaBSE, and E5 have been shown to perform robust cross-lingual retrieval in a multilingual environment \cite{chen2024bge}.

BGE-M3 is a recent advancement in the family of multilingual embedding models that supports over 100 languages and incorporates dense, sparse, and multi-vector-based retrieval models \cite{chen2024bge}. Its ability to process long text and simultaneously perform robust cross-lingual retrieval makes it a suitable candidate for Khmer text retrieval tasks. BGE-M3 demonstrated robust performance in a multilingual environment in the MIRACL benchmark, where it handled various diverse languages, including low-resource ones \cite{zhang2022miracl}. Nevertheless, it is important to note that even with a robust embedding model, document segmentation strategies are of equal importance in determining the effectiveness of a retrieval model \cite{qu2024semanticchunking, gunther2025latechunking}.

\subsection{Low-Resource Language Processing}

Khmer has been identified as a low-resource language, considering the lack of annotated corpora and linguistic tools available for the language \cite{buoy2020khmer, chea2015khmer}. Unlike Latin-based scripts, there is no whitespace specification for word boundaries, and the Khmer script follows an abugida system, consisting of complex combinations of consonants and vowels \cite{chea2015khmer}. Although there has been significant progress in the development of word segmentation and part-of-speech tagging for the Khmer language, achieving high accuracy results \cite{sry2022khmerreview}, there has been little research carried out on document-level chunking for retrieval-based applications.

Current trends in Southeast Asian language modeling and code-mixed text processing have been identified in recent research works \cite{wongso2025codemix, seahelm2025}. This indicates the increased emphasis being placed on low-resource languages. However, the chunking approaches for embedding-based retrieval in Khmer agricultural documents have not been extensively explored \cite{bhasa2023}.

\section{Methodology}
\subsection{Architecture Overview}
The proposed retrieval pipeline consists of six interconnected stages: data preparation, chunking, embedding, indexing, retrieval, and evaluation \cite{lewis2020rag, merola2025reconstructing}. Initially, documents are preprocessed by converting PDFs to text and then normalizing the text into Markdown format. The text is subsequently segmented using different chunking strategies, embedded with the BGE-M3 model \cite{chen2024bge}, and indexed using FAISS \cite{johnson2017faiss} for efficient nearest neighbor retrieval. Finally, retrieval performance is quantitatively evaluated using ground-truth question–answer pairs under a 5-fold cross-validation\cite{reimers2019sentencebert}.

\begin{figure}[t]
\centering
\makebox[\linewidth][c]{\includegraphics[width=1.09\linewidth, trim=0 6.5cm 0 2cm, clip]{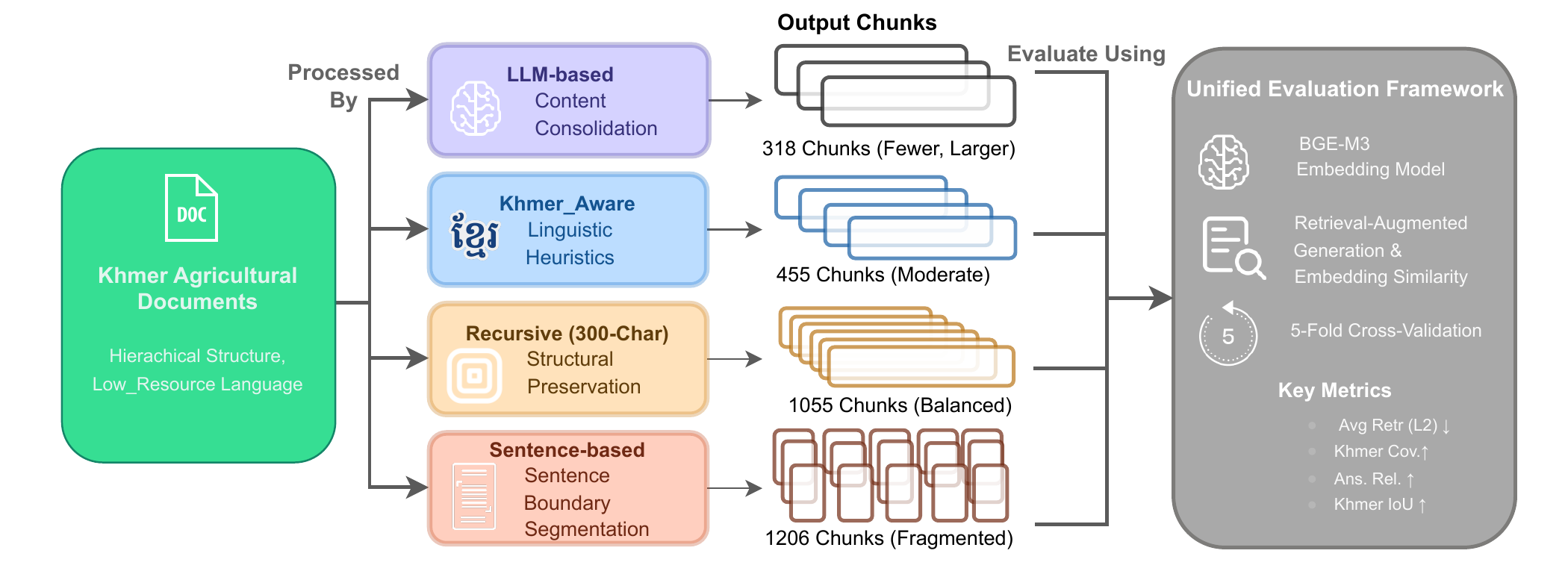}}
\caption{End-to-end pipeline chunking evaluation.}
\Description{Flow diagram of the retrieval pipeline with six stages: data preparation, chunking, embedding generation, vector indexing, retrieval, and evaluation.}
\label{fig:pipeline_overview}
\end{figure}

\subsection{Chunking Strategies}

Four chunking strategies are considered:

\begin{itemize}
    \item \textbf{Khmer-Aware Recursive Splitting:} A rule-based approach informed by Khmer script characteristics \cite{chea2015khmer, sry2022khmerreview}.
    \item \textbf{Recursive Splitting:} Rule-based segmentation using fixed character (e.g., 300 characters) to preserve hierarchical context \cite{koshorek2018textsegmentation}.
    \item \textbf{Sentence-Based Chunking:} Segmentation along sentence boundaries, designed to preserve semantic coherence at the sentence level \cite{reimers2019sentencebert}.
    \item \textbf{LLM-Based Semantic Chunking:} Large language model (LLM) guided segmentation using prompt-based semantic inference \cite{Strauss_2025}.
\end{itemize}

\subsubsection{Embedding Generation}

\textbf{BGE-M3 Model Selection and Configuration:}  
Each chunk is embedded into a dense vector using BGE-M3 \cite{chen2024bge}, a multilingual model effective for low-resource languages including Khmer. BGE-M3 produces 1,024-dimensional vectors for input sequences up to 8,192 tokens.

\subsubsection{Vector Indexing and Retrieval}

\textbf{FAISS Indexing and Similarity Search:}  
Embeddings are indexed with FAISS \cite{johnson2017faiss}, enabling fast nearest neighbor search. L2 distance is used to identify nearest chunks, while cosine similarity is applied to evaluate semantic relevance between retrieved chunks and queries \cite{reimers2019sentencebert}.

\subsubsection{Ground-Truth Construction}

\textbf{Question–Answer Pair Development:}  
Domain experts annotated the agricultural document collection to produce 18 question–answer pairs, which serve as ground truth for retrieval evaluation \cite{chea2015khmer, sry2022khmerreview}.

\subsection{Chunking Strategy Examples}

The following examples illustrate how different chunking strategies process Khmer agricultural text.

\begin{table}[t]
\centering
\caption{Comparison of Chunking Strategies with Examples}
\label{tab:chunking_examples}
\footnotesize
\begin{tabularx}{\linewidth}{@{}p{2.3cm} p{4.0cm} Y@{}}
\toprule
\textbf{Method} & \textbf{Description} & \textbf{Example} \\
\midrule

\textbf{Khmer-Aware} 
& Splits text hierarchically by paragraph (\texttt{\textbackslash n\textbackslash n}) and line breaks (\texttt{\textbackslash n}). Detects Khmer sentence markers such as ({\khmerfont ។ ៕}) to preserve boundaries.
& \textit{Input:} {\khmerfont ដាំស្វាយចន្ទី។ ថែរក្សាសួន៕} (Plant cashew trees. Maintain the garden.) \\
& &
Chunks: [{\khmerfont ដាំស្វាយចន្ទី។} (Plant cashew trees.)] \\
& &
\hspace{1.5cm}[{\khmerfont ថែរក្សាសួន៕} (Maintain the garden.)] \\
\midrule

\textbf{Recursive Standard}
& Hierarchical rule-based splitting using separators such as \texttt{\textbackslash n}, punctuation, and recursive chunking with overlap.
& \textit{Input:} Hello world. How are you? \\
& &
Chunks: [Hello world.] [How are you?] \\
\midrule

\textbf{Sentence-Based}
& Detects sentence boundaries using punctuation (e.g., {\khmerfont ។}) and groups sentences into chunks of fixed size ($N=5$) with overlap.
& \textit{Sentences:} S1–S8 \\
& &
Chunk 1: [S1–S5]; Chunk 2: [S5–S8] \\
\midrule

\textbf{LLM-Based}
& Uses a large language model to split text based on semantic coherence rather than fixed rules.
& \textit{Input:} Plant cashew trees. Water regularly. Protect crops. Spray pesticide. \\
& &
Chunks: [Plant cashew trees; Water regularly] \\
& &
\hspace{1.5cm}[Protect crops; Spray pesticide] \\
\bottomrule
\end{tabularx}
\end{table}

\section{Evaluation Metrics}

To present an overview of the proposed chunking methods—Recursive, Khmer-Aware Recursive, Sentence-Based, and LLM-Based four complementary evaluation metrics are adopted from recent studies on text chunking and retrieval~\cite{ gunther2025latechunking, Strauss_2025}. Evaluation is conducted using 5-fold cross-validation on 18 question–answer pairs, with results reported as mean ± standard deviation.

\subsection{Evaluation Setup }

To ground the metric definitions, consider the following representative example:

\begin{itemize}
    \item \textbf{Query ($Q$):} {\khmerfont តើត្រូវប្រើជីអ្វីខ្លះសម្រាប់ដើមស្វាយចន្ទី?} (What fertilizers should be used for cashew trees?)
    
    \item \textbf{Ground-truth Answer ($A$):} {\khmerfont គេត្រូវប្រើជីអ៊ុយរ៉េ និងជីប៉ូតាស្យូមសម្រាប់ដាំស្វាយចន្ទី។} (One should use Urea and Potassium fertilizers for planting cashews.)
    
    \item \textbf{Retrieved Chunks ($R$):}
    \begin{itemize}
        \item \textit{Chunk 1:} {\khmerfont ការប្រើប្រាស់ជីអ៊ុយរ៉េផ្តល់ផលល្អដល់ស្វាយចន្ទី។}(The use of Urea fertilizer gives good results to cashew nuts.)
        \item \textit{Chunk 2:} {\khmerfont ជីប៉ូតាស្យូមជួយឱ្យគ្រាប់ស្វាយចន្ទីមានទម្ងន់ធ្ងន់។}(Potassium fertilizer helps cashew nuts to be heavy.)
    \end{itemize}
\end{itemize}

\subsection{Semantic Alignment Metrics}

\subsubsection{Average Retrieval Score (L2 Distance)}

This metric measures the average L2 distance between the query embedding and the top-$k$ retrieved chunk embeddings in vector space~\cite{johnson2017faiss,reimers2019sentencebert}. Lower values indicate stronger semantic proximity.

\begin{equation}
    L2_{avg} = \frac{1}{k} \sum_{i=1}^{k} \| e_q - e_{c_i} \|_2
\end{equation}

\textit{Interpretation:} Smaller distances imply that retrieved chunks are closer to the query in embedding space.

\subsubsection{Answer Relevance}
Answer Relevance quantifies the semantic alignment between the aggregated top-$k$ retrieved chunks ($R$) and the ground-truth answer ($A$) using cosine similarity~\cite{reimers2019sentencebert}:

\begin{equation}
    AR = \text{cos}(e_R, e_A)
\end{equation}

where $e_R$ represents the mean embedding of the retrieved chunks.

Higher values indicate stronger semantic coverage of the expected answer.

\subsection{Khmer Linguistic Fidelity}

\subsubsection{Khmer Coverage}

To maintain script integrity and reduce unnecessary noise such as metadata, Latin script, or encoding artifacts, we measure the proportion of Khmer Unicode characters in the retrieved text. This metric is motivated by the challenges of segmentation in Khmer script, as discussed in previous studies on Khmer word segmentation and low-resource language processing~\cite{sry2022khmerreview}.

\begin{equation}
    \text{Coverage} = 
    \frac{\text{Total Khmer Characters in } R}
    {\text{Total Non-whitespace Characters in } R}
\end{equation}

Values closer to 1.0 indicate higher linguistic purity.

\subsubsection{Khmer Intersection-over-Union (KIoU)}

Standard word-level IoU is unreliable for Khmer due to the absence of explicit word boundaries and the presence of zero-width characters. To address this, we adopt a character-level IoU specifically adapted for Khmer. Unique Khmer character sets are extracted from both the retrieved text ($R$) and the ground-truth answer ($A$) to compute the intersection-over-union, providing a metric that is robust to segmentation inconsistencies in low-resource scripts~\cite{chea2015khmer}.

\begin{equation}
    KIoU = \frac{|R_{kh} \cap A_{kh}|}{|R_{kh} \cup A_{kh}|}
\end{equation}

This metric measures textual overlap while remaining robust to segmentation inconsistencies and script-specific characteristics.

\section{Evaluation Results}

In this section, we evaluate four distinct chunking strategies for a collection of Khmer agricultural documents: an LLM-based method, a Khmer-Aware Recursive method, a Recursive method with a 300-character limit, and a Sentence-based method. The results are presented as mean ± standard deviation across 5-fold cross-validation. The evaluation framework follows prior work on retrieval-augmented generation and embedding-based similarity measures~\cite{lewis2020rag}. Chunking strategies are inspired by previous studies in supervised text segmentation and long-context embedding~\cite{koshorek2018textsegmentation, qu2024semanticchunking}.

\begin{table}[t]
\footnotesize
\raggedright
\caption{Comparison of Chunking Strategies on Khmer Agricultural Documents}
\label{tab:evaluation_summary}
\setlength{\tabcolsep}{3.8pt}
\hspace*{-0.01\textwidth}\makebox[\textwidth][l]{
\begin{tabular}{lccccc}
\toprule
\textbf{Method} & \textbf{Chunks QTY} & \textbf{Avg Retr. (L2)↓} & \textbf{Khmer Cov.↑} & \textbf{Ans. Rel. (Cos)↑} & \textbf{Khmer IoU↑} \\
\midrule
LLM                & 318  & 0.4948 ± 0.0347 & 0.8737 ± 0.0184 & 0.8355 ± 0.0173 & 0.5766 ± 0.0141 \\
Khmer-Aware       & 455  & 0.7727 ± 0.0223 & 0.8425 ± 0.0389 & 0.8530 ± 0.0249 & 0.6053 ± 0.0269 \\
\textbf{Recursive} & 1055 & \textbf{0.4295 ± 0.0461} & \textbf{0.8860 ± 0.0231} & \textbf{0.8663 ± 0.0199} & \textbf{0.6441 ± 0.0347} \\
Sentence-based     & 1206 & 0.4721 ± 0.0470 & 0.8155 ± 0.0430 & 0.8556 ± 0.0230 & 0.6384 ± 0.0203 \\
\bottomrule
\end{tabular}
}
\textit{Note:} For Avg Retr. (L2)$\downarrow$, lower values are better; for Khmer Cov.$\uparrow$, Ans. Rel. (Cos)$\uparrow$, and Khmer IoU$\uparrow$, higher values are better.\par
\end{table}

\subsection{Chunk Distribution Analysis}

Chunk distribution differs across methods, reflecting each method's segmentation philosophy. The Sentence-based method produced the most chunks (1,206), resulting in fragmented procedural steps, consistent with prior observations~\cite{koshorek2018textsegmentation}. In contrast, the LLM-based method grouped content into fewer chunks (318), consolidating information but reducing granularity.

The Recursive method with a 300-character window produced 1,055 chunks, offering a balance between preserving hierarchical structure and capturing fine-grained semantic details. Compared to larger recursive windows (e.g., 500 or 800 characters), the 300-character improves embedding discrimination and retrieval performance~\cite{qu2024semanticchunking}.

\subsection{Retrieval Performance Comparison}

Table~\ref{tab:evaluation_summary} shows that the Recursive method performs best overall. It has the lowest L2 distance ($0.4295 \pm 0.0461$, where lower is better), and the highest Answer Relevance ($0.8663 \pm 0.0199$) and Khmer IoU ($0.6441 \pm 0.0347$, where higher is better)~\cite{lewis2020rag, guu2020realm}.

A paired t-test confirms that Recursive significantly outperforms Sentence-based chunking in L2 distance ($p = 0.0121$), indicating a statistically meaningful improvement.

Although Sentence-based chunking achieves competitive Khmer IoU ($0.6384 \pm 0.0203$), its retrieval distance ($0.4721 \pm 0.0470$) remains inferior to Recursive. The Khmer-Aware method shows high Khmer IoU but substantially worse L2 distance ($0.7727 \pm 0.0223$), suggesting that linguistic heuristics alone are insufficient without structural segmentation awareness.

\subsection{Qualitative Analysis}

Qualitative analysis shows that Recursive (300-character) is effective in preserving procedural integrity in Khmer agricultural guides. By retaining hierarchical structures such as headings, paragraphs, and lists, prerequisite conditions and corresponding actions remain semantically complete~\cite{koshorek2018textsegmentation, lewis2020rag}.

Sentence-based chunking often separates conditions from actions. Although it can match key terms, it sometimes lacks context, reducing coherence~\cite{reimers2019sentencebert, chea2015khmer}.

The LLM-based method produced moderate results but occasionally over-generalized domain-specific terminology, reducing alignment with queries for procedural instructions~\cite{chen2024bge}.

Overall, qualitative results indicate that the Recursive (300-character) method best balances contextual completeness and semantic specificity in Khmer agricultural documents~\cite{bhasa2023, sry2022khmerreview}.

\section{Discussion}

\subsection{Interpretation of Key Findings}
The results show that structural preservation has a more decisive effect than sentence-level segmentation in technical domains \cite{koshorek2018textsegmentation}. The Recursive (300-character) approach achieved the lowest L2 distance (0.4295 ± 0.0461) and highest Answer Relevance (0.8663 ± 0.0199) and Khmer IoU (0.6441 ± 0.0347). Moreover, a paired t-test confirmed that the improvement over Sentence-based chunking is statistically significant for L2 distance at p = 0.0121.

This achievement is due to the Recursive approach's ability to preserve “contextual bundles” \cite{gunther2025latechunking, qu2024semanticchunking}. In agricultural manuals, instructions are often given within a hierarchical structure of headers, paragraphs of explanations, and nested steps. The Recursive approach maintains this structural relationship within a 300-character window. As a result, the conditional statements are kept within a coherent embedding space with their corresponding actions \cite{chen2024bge}.

The Sentence-based approach achieved a comparable Khmer IoU of 0.6384 ± 0.0203 but has a disadvantage of breaking down procedural logic by disconnecting conditional statements from their corresponding actions \cite{reimers2019sentencebert}. The Khmer-Aware approach achieved a significantly higher L2 distance of 0.7727 ± 0.0223. Although it is a linguistically motivated approach, it is still far from a proper solution. The lack of segmentation constraints is a limitation of this approach, as emphasized by Chea \cite{chea2015khmer}.

\subsection{Trade-offs and Strategy Selection Guidelines}

The selection of an effective chunking strategy depends on a balance of four main factors: semantic preservation, linguistic integrity, computational efficiency, and structural preservation. These factors have been discussed in various studies on text segmentation and chunking \cite{koshorek2018textsegmentation, qu2024semanticchunking}.

\textbf{Khmer-Aware:}
This strategy be used when preserving Khmer punctuation markers and script-specific cues is important. It produces a moderate chunk volume of 455 chunks, but in this study it shows lower retrieval performance than the Recursive method.

\textbf{Recursive (300-character):} 
 This chunking strategy may be used in documents with a hierarchical structure in which the document structure reflects the semantic content of the text. It provides the highest retrieval accuracy with a reasonable index size of 1,055 chunks.

\textbf{LLM-Based:}
It provides the highest compression rate of 318 chunks. Hence, it may be used in applications in which storage space is a concern.

\textbf{Sentence-Based:}
 It used in documents with a linguistic structure in which sentence boundaries play a crucial role in the content. It provides the lowest retrieval accuracy with a chunk volume of 1,206 chunks.

The results show that the granularity of the segment may be determined based on the document structure rather than linguistic factors.

\subsection{Implications for Low-Resource Language Processing}

The results show a guideline for designing retrieval systems in low-resource language documents. It indicates that preserving the document structure may compensate for linguistic chunking errors \cite{chea2015khmer}.

The results may be used in designing retrieval systems in Khmer documents in which word boundaries are not clearly defined. Sentence tokenization may not be effective in Khmer documents. Hence, preserving the document structure may be used in designing retrieval systems.

The results may be used in designing retrieval systems in documents written in various Southeast Asian scripts in which word boundaries are not clearly defined. It may be used in designing retrieval systems in Lao and Thai documents \cite{seahelm2025}.

\subsection{Limitations}

The research has some limitations. Firstly, the research only focuses on the evaluation of Khmer agricultural manuals. It may not be generalized to conversational, literary, or multi-domain corpora \cite{lewis2020rag}. Secondly, the experiments only used one multilingual embedding model (BGE-M3) \cite{chen2024bge}. Other embedding architectures may have different sensitivity to chunk length and structural segmentation. Thirdly, the LLM-based method was only evaluated with one model configuration and prompt design. It may have a higher performance after optimization \cite{liu2023lostmiddle}. Lastly, although using objective quantitative metrics in the research helps in the objective comparison of results, using human evaluation in the agricultural domain would have increased the validity of the research in assessing the quality of retrieval \cite{seahelm2025}.

\section{Conclusion and Future Work}

This paper provides a comprehensive evaluation of text chunking strategies in embedding-based text retrieval in a low-resource language setting, Khmer agricultural documents \cite{sry2022khmerreview}. A unified evaluation framework is proposed to incorporate semantic-based text retrieval metrics (L2 distance and cosine similarity) and language model-based fidelity measures (Khmer Coverage and Khmer IoU) \cite{lewis2020rag}. Performance is evaluated using 5-fold cross-validation to ensure statistical validity. Experimental results demonstrate that structure-preserving recursive chunking (300-character setting) provides the most promising results in text retrieval in a low-resource language setting, achieving minimal text retrieval distance and maximum semantic similarity, with statistically significant improvements over Sentence-based segmentation \cite{gunther2025latechunking}.

These results confirm that text chunking strategies significantly impact text retrieval performance in a low-resource language setting \cite{seahelm2025, bhasa2023}. Document structure and contextual continuity are key factors in text embeddings that provide more discriminative text embeddings compared to sentence-based or script-based text segmentation \cite{reimers2019sentencebert, chea2015khmer}. Although text chunking using LLMs provides compact text embeddings and adaptive topic boundaries \cite{liu2023lostmiddle}, it is not as effective as structure-preserving recursive splitting in text retrieval in a technical procedural document setting. In summary, it is believed that document structure is a more important factor in text retrieval performance compared to text chunking strategies in a low-resource language setting \cite{sry2022khmerreview}.

Possible future work in this area is to explore hybrid text segmentation strategies that integrate structure-based recursive splitting and semantic text refinement \cite{qu2024semanticchunking}. More text embeddings and reranking strategies will be investigated to assess text chunking granularity sensitivity \cite{chen2024bge, lewis2020rag}. The proposed text segmentation and evaluation framework will also be extended to explore text retrieval in other low-resource language and in other domains beyond agricultural documents.


\bibliographystyle{IEEEtran}
\bibliography{base}

\end{document}